\documentclass{article}

\usepackage[utf8]{inputenc} % allow utf-8 input
\usepackage[T1]{fontenc}    % use 8-bit T1 fonts
\usepackage{arxiv}

\usepackage{booktabs}       % professional-quality tables
\usepackage{amsfonts}       % blackboard math symbols
\usepackage{nicefrac}       % compact symbols for 1/2, etc.
\usepackage{microtype}      % microtypography
\usepackage{xcolor} % colors
\usepackage{graphicx}
\usepackage{amsmath} 
\usepackage{natbib}  
\usepackage{placeins}
\usepackage{hyperref}

\title{Single Molecule Localization Microscopy Challenge: \\ A Biologically Inspired
Benchmark \\ for Long-Sequence Modeling}

\author{
  Fatemeh Valeh \\
  Department of Biophysics\\
  Technische Universit\"at Wien\\
  Vienna, Austria \\
  \texttt{fatemeh.valeh@tuwien.ac.at}
  \And
  Monika Farsang \\
  Department of Informatics\\
  Technische Universit\"at Wien\\
  Vienna, Austria  \\
  \texttt{monika.farsang@tuwien.ac.at}
  \And
  Radu Grosu \\
  Department of Informatics\\
  Technische Universit\"at Wien\\
  Vienna, Austria  \\
  \texttt{radu.grosu@tuwien.ac.at}
  \And
  Gerhard Schütz\\
  Department of Biophysics\\
  Technische Universit\"at Wien\\
  Vienna, Austria  \\
  \texttt{gerhard.schuetz@tuwien.ac.at}
}

\begin{document}

\maketitle
\begin{abstract}
State space models (SSMs) have recently achieved strong performance on long-sequence modeling tasks while offering improved memory and computational efficiency compared to transformer-based architectures. However, their evaluation has been largely limited to synthetic benchmarks and application domains such as language and audio, leaving their behavior on sparse and stochastic temporal processes in biological imaging unexplored. In this work, we introduce the Single Molecule Localization Microscopy Challenge (SMLM-C), a benchmark dataset consisting of ten SMLM simulations—spanning dSTORM and DNA-PAINT modalities with varying hyperparameter—designed to evaluate state-space models on biologically realistic spatiotemporal point-process data with known ground truth. Using a controlled subset of these simulations, we evaluate state space models and find that performance degrades substantially as temporal discontinuity increases, revealing fundamental challenges in modeling heavy-tailed blinking dynamics. These results highlight the need for sequence models better suited to sparse, irregular temporal processes encountered in real-world scientific imaging data.
\end{abstract}

\section{Introduction}

State space models (SSMs), including structured variants such as S4 \cite{gu2021efficiently}, S5 \cite{smith2022simplified}, and selective architectures such as Mamba \cite{gu2024mamba}, have recently demonstrated strong performance on long-sequence modeling tasks. By offering improved memory efficiency and favorable scaling compared to transformer-based models, these architectures have shown competitive results across language, audio, and vision benchmarks. However, evaluation of SSMs has largely focused on synthetic tasks or domains characterized by dense, regularly sampled, and relatively smooth temporal signals. Their behavior under biologically realistic conditions—where temporal dynamics may be sparse, irregular, and heavy-tailed—remains largely unexplored.

Single Molecule Localization Microscopy (SMLM) provides a compelling testbed for such conditions. SMLM comprises a family of super-resolution fluorescence microscopy techniques that overcome the diffraction limit by temporally separating the activation of individual fluorophores \cite{lelek2021single}. Rather than imaging all emitters simultaneously, only sparse subsets enter an emissive (“on”) state in any given frame. Each active fluorophore is localized with nanometer precision by fitting its point spread function, and a high-resolution image is reconstructed by aggregating these localizations over thousands of frames.

The temporal signals underlying SMLM differ fundamentally from standard long-sequence benchmarks. Individual fluorophores stochastically transition between emissive (“on”) and non-emissive (“off”) states, producing sparse and highly discontinuous localization sequences that can span thousands of frames. These blinking dynamics are often heavy-tailed \cite{platzer2020unscrambling}, with prolonged periods of inactivity punctuated by brief emission bursts. Experimental data are further corrupted by photon shot noise, optical aberrations, detector noise, and algorithmic filtering applied by standard reconstruction pipelines \cite{ovesny2014thunderstorm}.

Together, these properties yield sequences that are sparse, irregular, non-stationary, and noise-corrupted—substantially departing from the dense and stationary signals on which long-context sequence models are typically evaluated. Modeling such data requires handling heavy-tailed dwell-time distributions, maintaining long-range memory across extended inactive intervals, and remaining robust to measurement uncertainty. These challenges position SMLM as a natural and demanding benchmark for modern SSM architectures.

Despite the central role of simulation in SMLM research, existing benchmarks do not target long-context sequence modeling. Because true emitter positions are unobservable in experimental data, evaluation relies heavily on simulated datasets with known ground truth \citep{sage2019super}. Community challenges primarily assess frame-level localization accuracy or reconstructed image quality \cite{sage2015quantitative,sage2019super}. They are not designed to evaluate models that integrate information across entire acquisition sequences to suppress blinking-induced artifacts.

Recent deep learning approaches for SMLM reconstruction likewise rely on simulated data \citep{nehme2018deep,ouyang2018deep,speiser2021deep}, but predominantly employ convolutional architectures operating on individual frames or short temporal windows. To our knowledge, no existing benchmark explicitly evaluates long-context sequence models designed to exploit extended temporal dependencies in blinking dynamics. Related efforts in biological imaging, such as the Cell Tracking Challenge \citep{ulman2017objective}, focus on densely annotated microscopy videos with regular sampling and predominantly local temporal dependencies, lacking the extreme sparsity and heavy-tailed behavior characteristic of SMLM data.

This gap motivates the development of a dedicated benchmark for evaluating long-sequence models under biologically realistic spatiotemporal sparsity. To address this need, we introduce the Single Molecule Localization Microscopy Challenge (SMLM-C), a simulation-based benchmark specifically designed to evaluate sequential models on sparse localization data with complete ground-truth emitter positions.

SMLM-C is constructed using a simulation engine that explicitly models fluorophore blinking kinetics, emitter density variation, per-frame localization uncertainty arising from the imaging process, and algorithmic detection constraints. The benchmark comprises ten experimentally motivated scenarios spanning dSTORM and DNA-PAINT imaging modalities, with sequences extending up to 10,000 frames. Each dataset forms a sparse spatiotemporal point process whose statistics cannot be captured by simple stationary or short-memory models. Even in nominally low-density regimes, random spatial arrangements can produce overlapping localization patterns, requiring models to disambiguate intermittent emissions by jointly exploiting spatial and temporal context.

Using a controlled subset of dSTORM simulations from SMLM-C, we evaluate two representative state space models—S5 and Mamba—on the task of predicting ground-truth emitter positions from observed localization sequences. By varying the average off-time between emission events under otherwise fixed conditions, we isolate the impact of temporal discontinuity on model performance.

Our empirical results reveal significant limitations in current long-context sequence models. While both structured and selective state space architectures demonstrate some capacity to aggregate information over time, all evaluated models struggle under the combined effects of extreme sparsity, heavy-tailed blinking, and realistic detection noise. These findings highlight the need for further methodological advances in sequence modeling to address the challenges posed by physically grounded biological data.

Contributions.
\begin{itemize}
\item We introduce SMLM-C, a biologically inspired simulation benchmark for evaluating long-sequence models on sparse spatiotemporal localization data with known ground truth.
\item We design simulation regimes that capture key challenges of SMLM imaging, including temporal sparsity, heavy-tailed blinking dynamics, and realistic localization noise.
\item We perform a controlled empirical evaluation of modern state space models, isolating the impact of increasing temporal discontinuity on localization performance.
\end{itemize}

\section{Dataset}
We introduce the Single Molecule Localization Microscopy Challenge (SMLM-C), a benchmark of simulated single-molecule localization sequences designed to evaluate long-context sequence models on sparse, stochastic temporal point processes with known ground truth. Across ten experimental conditions, SMLM-C varies imaging modality (dSTORM and DNA-PAINT), emitter density, and blinking kinetics, producing sequences up to 10{,}000 frames.

Due to the computational cost of training long-sequence models, we focus in this work on two representative dSTORM conditions, D2 and D4 (see Table~\ref{tab:dataset_conditions}), which isolate the effect of temporal discontinuity. These conditions differ only in the average off-time between emission events ($\mu_{\mathrm{off}}=100$ vs.\ $1000$ frames), while holding emitter density and the expected number of localizations per emitter fixed. Full simulation details are deferred to the appendix.

\begin{table}[b]
\caption{Overview of simulated experimental conditions in SMLM-C.}
\label{tab:dataset_conditions}
\centering
\small
\begin{tabular}{llccccc}
\toprule
Condition & Modality & Density & $\mu_{\mathrm{on}}$ & $\mu_{\mathrm{off}}$ & Frames & Blinking \\
 &  & (molec./$\mu$m$^2$) & (frames) & (frames) &  & termination model \\
\midrule
D1 & dSTORM & 50 & 5 & 100  & 6305 & Exponential ($\mu=20$) \\
D2 & dSTORM & 50 & 5 & 100  & 10{,}000 & Exponential ($\mu=50$) \\
D3 & dSTORM & 50 & 5 & 1000 & 10{,}000 & Exponential ($\mu=20$) \\
D4 & dSTORM & 50 & 5 & 1000 & 10{,}000 & Exponential ($\mu=50$) \\
D5 & dSTORM & 1000 & 5 & 1000 & 10{,}000 & Exponential ($\mu=20$) \\
D6 & dSTORM & 1000 & 5 & 1000 & 10{,}000 & Exponential ($\mu=50$) \\
\midrule
P1 & DNA-PAINT & 50 & 5 & 100  & 4583 &  Poisson ($\lambda=50$) \\
P2 & DNA-PAINT & 50 & 5 & 100 & 10{,}000 & Unlimited \\
P3 & DNA-PAINT & 50 & 5 & 1000 & 10{,}000 & Unlimited \\
P4 & DNA-PAINT & 1000 & 5 & 1000  & 10{,}000 & Unlimited \\
\bottomrule
\end{tabular}

\vspace{3pt}
\footnotesize{
For dSTORM, $\mu$ denotes the mean of the exponential distribution governing the number of localization events per emitter.
}
\end{table}

\section{Methods}
\subsection{Model architectures}

We evaluate two state-of-the-art sequence models designed for long-context modeling:

\paragraph{S5.}~\citep{smith2022simplified}:
A simplified state space model that employs diagonal state matrices and parallel scan operations for efficient training.
We consider two configurations (small and large) with hidden dimensions $d \in \{128, 256\}$ and corresponding SSM state dimensions
$N \in \{256, 512\}$, each using 6 layers. We denote them as S5-S and S5-L, respectively.

\paragraph{Mamba2.}~\citep{dao2024transformers}:
A selective state space model with input-dependent state transitions.
We evaluate small and large variants with hidden dimensions $d \in \{128, 256\}$ and a fixed state dimension $N=64$, using 6 layers. We denote them as Mamba-2-S and Mamba-2-L, respectively.

\paragraph{Regression head.}
For both S5 and Mamba-2, we use the same lightweight MLP decoder to map the pooled representation $\mathbf{h}$ to $N$ emitter coordinates.
Specifically, the decoder is a 3-layer MLP that takes as input a pooled representation of dimensionality $d$, where $d$ denotes the hidden dimensionality of the underlying SSM encoder. The representation is passed through a hidden layer of size $d$ and a second hidden layer of size $d/2$, with GELU nonlinearities in between. The final linear layer outputs the values of $2N$ followed by a sigmoid, and the result is reshaped to $(N,2)$ to represent the coordinates of the emitters. Coordinates are trained in normalized units and converted back to nanometers for evaluation.

\paragraph{Implementation.}
S5 models are implemented in JAX/Flax and Mamba-2 models in PyTorch, following their official implementations. To ensure a fair comparison, we use an identical regression head and pooling scheme across both model families.

\subsection{Task formulation}

We formulate SMLM reconstruction as a \emph{sequence-to-set} prediction task aimed at removing blinking-induced artifacts and recovering
the true underlying emitter locations.
Given a sequence of observed localizations
$\mathbf{X} = \{(x_t, y_t, t)\}_{t=1}^T$,
where each observation consists of spatial coordinates and a frame index,
the goal is to predict the ground-truth emitter positions
$\mathbf{Y} = \{(\hat{x}_i, \hat{y}_i)\}_{i=1}^N$
corresponding to the $N$ physical emitters present in the field of view.

Multiple localization events may originate from the same emitter due to stochastic blinking, and repeated localizations from a single
emitter are subject to measurement noise, appearing at slightly different spatial positions modeled as isotropic Gaussian noise with
standard deviation $\sigma = 10$~nm.
When emitters are spatially proximal, the resulting localization clouds may significantly overlap, making it ambiguous or even
impossible to assign observations to emitters using spatial information alone.
Accurate reconstruction in such cases therefore requires jointly exploiting spatial and temporal cues to disentangle blinking events
and recover the true emitter positions.
In our simulations, the number of physical emitters $N$ is fixed within each experimental condition.
Emitter positions are initially sampled according to the specified density.
For the low-density dSTORM conditions (50 molecules/$\mu$m$^2$) and a $500 \times 500$~nm region of interest (area $0.25~\mu$m$^2$), this corresponds to an expected count of 12.5 emitters, which we round to 12 initial emitters per simulation.

During simulation, a detection filtering step is applied that removes localizations that violate the per-frame detection constraints.
As a result, some emitters may produce no retained localizations over the full acquisition and are therefore excluded from the observed dataset.
Consequently, the number of emitters with at least one detected localization is smaller than the number initially placed.

For the dSTORM conditions evaluated in this work, this process results in $N=7$ retained emitters for condition D2 and $N=9$ retained emitters for condition D4.
Because the simulation is stochastic, the exact number of retained emitters can vary slightly between runs.
To ensure a consistent output dimensionality for sequence-to-set prediction, we fix $N$ to the minimum number of retained emitters observed in all simulations within each condition.
We therefore treat $N$ as known and do not address the estimation of the emitter count in this work.
Accordingly, the model output is a fixed-size, permutation-invariant set of emitter positions.

Because the region of interst is $500 \times 500$~nm and detection filtering threshold is also set to 500~nm, at most one localization event is retained per frame in this generated dataset. We note that for larger regions of interest or relaxed filtering thresholds, multiple localizations per frame may occur, which our formulation naturally supports.
Input sequences are represented as sparse frame-wise tensors, where empty frames are padded with dummy values $(0,0)$ and masked
during model computation to prevent them from contributing to the loss.
Due to stochastic blinking and photobleaching, total sequence length may vary across realizations within the same condition;
all sequences are therefore padded to the maximum sequence length observed within each simulation condition.

\paragraph{Input representation and masking.}

Each frame is represented by a 2D coordinate $(x_t,y_t)$, with empty frames padded and masked. Models process the full padded sequence, and a masked mean pooling operation over time produces a single sequence-level representation of dimension $d$.
Overall, our formulation corresponds to a multi-input, single-output prediction task: a variable-length sequence of localization observations is mapped to a single output, namely a fixed-size, permutation-invariant set of emitter coordinates.

\subsection{Training details}

We follow the default training recipes for each model family and only modify the components required by our regression task.
Specifically, we replace the classification head with an MLP decoder that predicts $N \times 2$ coordinates corresponding to the $(x,y)$ positions of the $N$ emitters.
All other architectural and optimizer hyperparameters are kept at their reference defaults for S5 and Mamba-2.
As a result, some training settings (e.g., learning rate schedules) differ between the two model families.

All models are trained using the AdamW optimizer to minimize a Chamfer distance loss between predicted and ground-truth emitter sets \citep{fan2017point}.
We use the unsquared Euclidean variant, which computes average nearest-neighbor distances in both directions.

We use a batch size of 64 and train for up to 100 epochs, selecting the checkpoint with the lowest validation Hungarian error.
Experiments are repeated with three random seeds and we report mean $\pm$ standard deviation.
All reported results use the same evaluation protocol and model selection criterion across architectures.
All experiments were run on a single NVIDIA A100 GPU.

\subsection{Evaluation metrics}

Model performance on the test set is evaluated using detection-style metrics and localization precision measures designed to separately quantify emitter recovery and spatial accuracy.
While Hungarian error is used for model selection, final test performance is reported using detection accuracy, TP/FP/FN counts at fixed spatial thresholds, and RMSE on true-positive matches.

\paragraph{Hungarian error (model selection).}
For model selection, we utilize the Hungarian matching error, which computes an optimal one-to-one assignment between predicted and ground-truth emitter positions by minimizing the total Euclidean distance between matched pairs \citep{kuhn1955hungarian}. The Hungarian error is defined as the mean Euclidean distance of the matched pairs, providing a smooth, permutation-invariant measure of global localization error. Although we do not use the Hungarian error for training due to its high computational cost, it is more suitable for our task as it enforces a one-to-one assignment. In contrast, the Chamfer distance allows for multiple assignments to a single ground truth, which may not align with our objectives.

\paragraph{Detection metrics and localization precision (test evaluation).}
For test evaluation, we first establish a one-to-one correspondence between predicted and ground-truth emitters using the Hungarian algorithm.
Matched pairs whose distance is below a fixed threshold $\tau$ are counted as true positives (TP).
Matched pairs with distance exceeding $\tau$ are counted as  false positive (FP)  corresponding a missed emitter
Since the model is constrained to predict exactly $N$ emitter locations per sample—where $N$ is the known number of physical emitters in the simulation—the numbers of FP and false negatives (FN) are always equal and are therefore reported jointly. Whenever the algorithm misses one true emitter location, it will result in a false prediction elsewhere.
Detection accuracy is defined as $\mathrm{TP}/N$, measuring the fraction of correctly recovered emitters.
To quantify localization precision independently of detection failures, we compute the root mean squared error over true-positive matches only, denoted $\mathrm{RMSE}_{\mathrm{TP}}$.
All distances are measured in nanometers.

\paragraph{Evaluation protocol.}
All metrics are computed on the held-out test split, which contains 10{,}000 samples per dataset.
Each simulation condition consists of 100{,}000 samples in total, split into 80\% training, 10\% validation, and 10\% test sets.
Final performance is reported exclusively on the test set.
All experiments are repeated with three random seeds, and results are reported as mean $\pm$ standard deviation.

\section{Results}

We evaluate S5 and Mamba-2 models on two representative dSTORM simulation conditions that differ primarily in the average
off-time between emission events, corresponding to short and long dark-state regimes.

Table~\ref{tab:dstorm_main_hungarian} reports validation  and test Chamfer loss and Hungarian error for all model variants.

\begin{table}[t]
\caption{Validation and test performance on dSTORM simulations. Results are mean $\pm$ std over three seeds. Models are selected by validation Hungarian error.}
\centering
\small
\setlength{\tabcolsep}{6pt}
\begin{tabular}{lcccc}
\toprule
 & \multicolumn{4}{c}{\textbf{dSTORM ($\mu_{\mathrm{off}} = 100 \mathrm{~frames}$)}} \\
\cmidrule(lr){2-5}
 & S5-S & S5-L & Mamba-2-S & Mamba-2-L \\
\midrule
Val loss  & 0.0445$\pm$0.0007 & 0.0396$\pm$0.0015 & 0.0515$\pm$0.0044 & 0.0435$\pm$0.0009 \\
Test loss & 0.0442$\pm$0.0004 & 0.0392$\pm$0.0013 & 0.0513$\pm$0.0040 & 0.0435$\pm$0.0009 \\
\midrule
Val Hungarian (nm)  & 34.50$\pm$0.61 & 33.24$\pm$2.43 & 39.09$\pm$4.24 & 41.32$\pm$0.73 \\
Test Hungarian (nm) & 34.16$\pm$0.52 & 32.95$\pm$2.45 & 38.79$\pm$4.14 & 41.04$\pm$0.53 \\
\midrule\midrule
 & \multicolumn{4}{c}{\textbf{dSTORM ($\mu_{\mathrm{off}} = 1000\mathrm{~frames}$)}} \\
\cmidrule(lr){2-5}
 & S5-S & S5-L & Mamba-2-S & Mamba-2-L \\
\midrule
Val loss  & 0.0521$\pm$0.0044 & 0.0423$\pm$0.0012 & 0.0520$\pm$0.0027 & 0.0391$\pm$0.0013 \\
Test loss & 0.0520$\pm$0.0043 & 0.0424$\pm$0.0012 & 0.0520$\pm$0.0028 & 0.0391$\pm$0.0011 \\
\midrule
Val Hungarian (nm)  & 43.06$\pm$1.17 & 37.93$\pm$0.74 & 39.36$\pm$0.80 & 35.46$\pm$0.64 \\
Test Hungarian (nm) & 42.92$\pm$1.46 & 38.18$\pm$0.73 & 39.46$\pm$0.71 & 35.53$\pm$0.49 \\
\bottomrule
\end{tabular}

\label{tab:dstorm_main_hungarian}
\end{table}

Across both datasets, larger model variants consistently achieve lower validation error, indicating an improved capacity for modeling long-range temporal dependencies induced by stochastic blinking. Notably, the S5 variant outperformed Mamba-2 on the dSTORM D2 dataset, which features a shorter dark state. In contrast, Mamba-2 demonstrated better performance on the dSTORM D4 dataset, characterized by a longer dark state.

Table~\ref{tab:dstorm_main_detection} summarizes the evaluation of final metrics on test dataset. The findings from this table are also consistent with those in the previous table.

\begin{table}[t]
\caption{Detection quality and localization precision on dSTORM simulations at a 20 nm matching threshold. Results are mean $\pm$ std over three seeds.}
\centering
\small
\setlength{\tabcolsep}{6pt}
\begin{tabular}{lcccc}
\toprule
 & \multicolumn{4}{c}{\textbf{Model}} \\
\cmidrule(lr){2-5}
Metric & S5-S & S5-L & Mamba-2-S & Mamba-2-L \\
\midrule
% \multicolumn{5}{l}{\textbf{dSTORM (}$\mu_{\mathrm{off}}=100\mathrm{~frames}$\textbf{), $N_{\text{emit}}=7$, $\tau=20$ nm}} \\
\multicolumn{5}{l}{\textbf{dSTORM (}$\mathbf{\mu_{\mathrm{off}}=100\ {frames}}$\textbf{), $\mathbf{N_{\text{emit}}=7}$, $\mathbf{\tau=20}$ nm}} \\

TP (count)             & 5.0507$\pm$0.0344 & 5.1378$\pm$0.0861 & 4.7548$\pm$0.2060 & 4.7401$\pm$0.0186 \\
FP/FN (count)          & 1.9493$\pm$0.0344 & 1.8622$\pm$0.0861 & 2.2452$\pm$0.2060 & 2.2599$\pm$0.0186 \\
Detection Acc. (\%) & 0.7215$\pm$0.0049 & 0.7340$\pm$0.0123 & 0.6793$\pm$0.0295 & 0.6772$\pm$0.0027 \\
RMSE$_{\mathrm{TP}}$ (nm)
                    & 6.0119$\pm$0.1382 & 5.4046$\pm$0.0876 & 6.7145$\pm$0.2163 & 5.2621$\pm$0.0603 \\
\midrule
\multicolumn{5}{l}{\textbf{dSTORM (}$\mathbf{\mu_{\mathrm{off}}=1000\ {frames}}$\textbf{), $\mathbf{N_{\text{emit}}=9}$, $\mathbf{\tau=20}$ nm}} \\

TP (count)             & 5.5571$\pm$0.1427 & 6.0054$\pm$0.0452 & 5.8026$\pm$0.0868 & 6.2626$\pm$0.0189 \\
FP/FN (count)          & 3.4429$\pm$0.1427 & 2.9946$\pm$0.0452 & 3.1974$\pm$0.0868 & 2.7374$\pm$0.0189 \\
Detection Acc. (\%) & 0.6175$\pm$0.0159 & 0.6673$\pm$0.0050 & 0.6447$\pm$0.0097 & 0.6959$\pm$0.0021 \\
RMSE$_{\mathrm{TP}}$ (nm)
                    & 6.6512$\pm$0.0896 & 5.8476$\pm$0.0833 & 7.2756$\pm$0.2403 & 5.7881$\pm$0.1275 \\
\bottomrule
\end{tabular}

\label{tab:dstorm_main_detection}
\end{table}

Both architectures achieve low localization error in the short off-time regime. However, performance degrades noticeably in the long off-time condition, where temporal sparsity increases and repeated blinking events are separated by extended dark periods. In this setting, Mamba-2 consistently outperforms S5, suggesting improved robustness to long-range temporal gaps.

\begin{figure}[t]
    \centering
    \includegraphics[width=\linewidth]{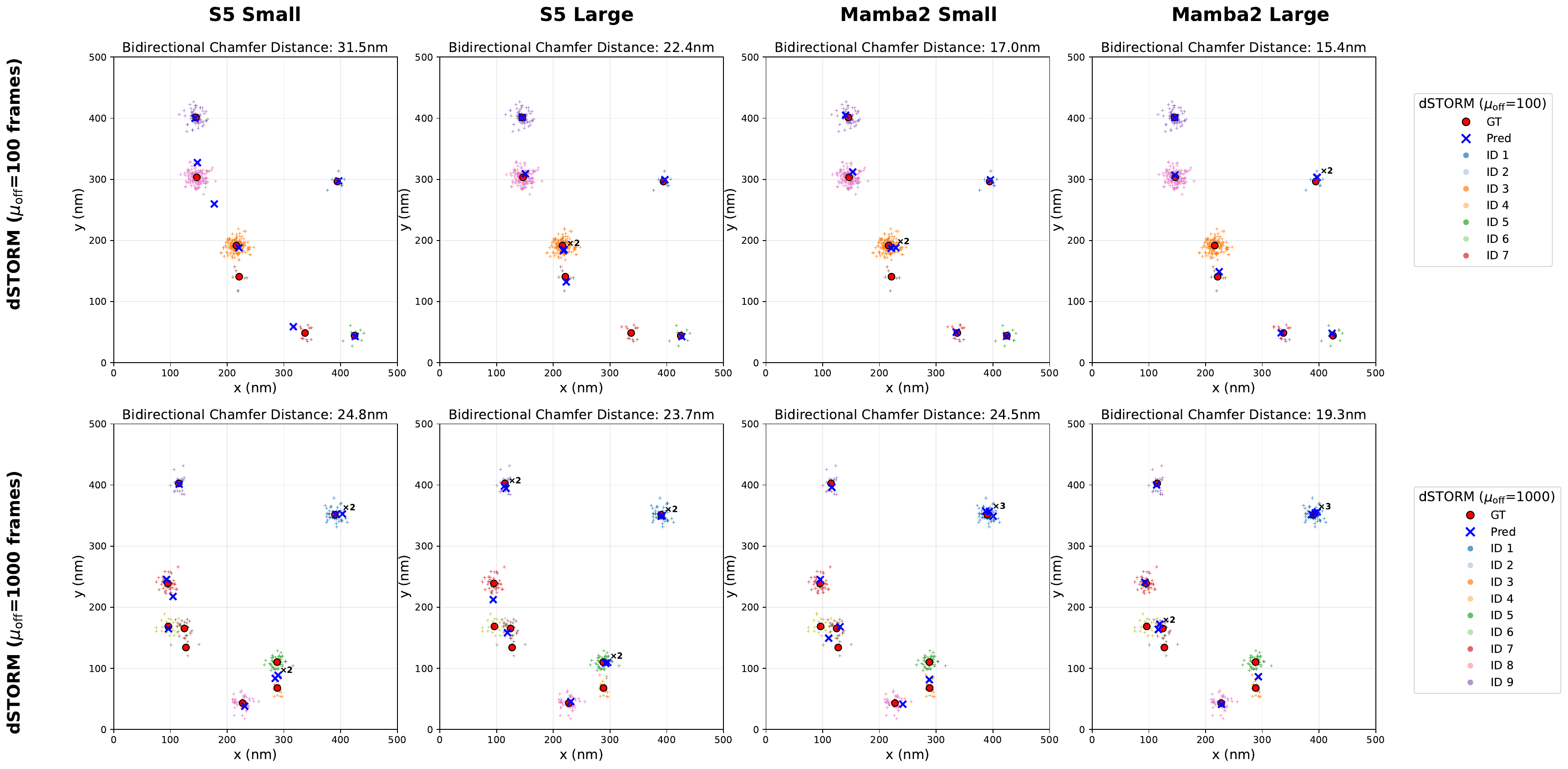}
    \caption{
    \textbf{Qualitative median-case comparison on dSTORM simulations.}
    Ground-truth emitter positions (red $\circ$) and model predictions (blue $\times$) 
    for representative median-difficulty samples from dSTORM with short off-time 
    ($\mu_{\text{off}} = 100$, top row) and long off-time ($\mu_{\text{off}} = 1000$, 
    bottom row). Colored background points show the raw observed localizations, where each color corresponds to a distinct emitter identity (ID).
    Columns correspond to S5 Small, S5 Large, Mamba-2 Small, and Mamba-2 Large. 
    Annotations $\times k$ indicate $k$ predicted emitter locations lying within 20~nm 
    of each other, and are placed for improved visibility.
    }
    \label{fig:median_case}
\end{figure}

To select representative qualitative examples, we first compute the Chamfer loss for every test sequence and each of the four model configurations (S5-S, S5-L, Mamba2-S, Mamba2-L). For each test sequence, losses are averaged across models to obtain a model-agnostic difficulty score. The median of this distribution is then computed, and the first test sequence whose averaged loss is closest to this median is selected as the representative ``median'' example shown in the main text  (Figure~\ref{fig:median_case}). This procedure avoids cherry-picking and ensures that the visualization reflects typical reconstruction performance.

\subsection{Detection quality and precision}

To further analyze detection behavior, we report true positives (TP), false positives / false negatives (FP/FN), and localization
precision on correctly matched emitters.
Results are summarized in Table~\ref{tab:dstorm_main_detection}.

While all models produce a fixed number of predictions per sample, differences emerge in the proportion of predictions that fall
within the true-positive threshold.
Larger models consistently achieve higher TP rates and lower RMSE on true positives, indicating more precise localization of
correctly identified emitters.

\section{Discussion}
Our results demonstrate that while long-context state space models can capture temporal dependencies in sparse SMLM localization sequences, their absolute reconstruction performance remains limited. The highest detection accuracy achieved was 73\%, indicating substantial room for improvement in recovering true emitter locations from temporally fragmented observations.

Despite this modest absolute performance, our experiments reveal several important insights into the fundamental challenges of SMLM reconstruction and the capabilities of modern sequence models. First, performance is strongly influenced by the temporal statistics of emitter blinking, with longer off-time regimes posing a substantially greater challenge across all models. This degradation as temporal sparsity increases underscores a fundamental difficulty: when blinking events are infrequent and spatially overlapping, reliable localization requires integrating information over extended time horizons---a task that remains challenging even for advanced architectures.

Second, both S5 and Mamba-2 demonstrate clear benefits from increased model capacity, with larger configurations consistently outperforming their smaller counterparts. This scaling behavior suggests that the models are indeed learning meaningful temporal representations, but require greater expressivity to disentangle repeated blinking events effectively. The capacity-performance relationship indicates that architectural advances in long-context modeling may translate into improved robustness under sparse temporal conditions, even if current absolute accuracy remains limited.

Third, Mamba-2 consistently outperforms S5 in the long off-time regime, suggesting that its input-dependent state transitions provide an advantage when relevant observations are separated by extended temporal gaps. However, this performance gain comes at a computational cost: Mamba-2 models train $2$--$3\times$ slower than their S5 counterparts at comparable scales and contain 29--40\% more parameters. The S5-L configuration contains approximately $2.08\,\text{M}$ parameters compared to $2.69\,\text{M}$ for Mamba-2-L, while training times are $14.1\,\text{h}$ versus $42.6\,\text{h}$ respectively (averaged over three independent runs).

Several limitations constrain the interpretation and applicability of our findings. First, we assume the number of emitters is known and do not address emitter count estimation---a critical prerequisite for practical deployment. Second, our experiments focus on a compact region of interest with at most one localization per frame. Extending this framework to larger fields of view and higher per-frame densities represents an essential direction for future work. Third, while our dataset includes a broader range of simulated conditions, we restrict evaluation to two representative dSTORM settings for computational efficiency, which may limit generalizability to other SMLM imaging modalities.

Additionally, exact parameter matching across model families was not feasible due to the use of standard architectural configurations. Performance differences therefore reflect both architectural characteristics and modest discrepancies in parameter count. Finally, and most importantly, the overall detection accuracy of 73\% falls short of what would be required for practical application in SMLM reconstruction pipelines.

The limited absolute performance suggests that set-based reconstruction using sequence models alone may not be sufficient for this task. Future work should investigate hybrid approaches that combine the temporal modeling capabilities demonstrated here with spatial priors, physical constraints, or complementary localization methods. The clear scaling trends and ability to capture blinking dynamics indicate that state space models provide a useful foundation, but significant architectural innovation or task reformulation will be necessary to achieve performance competitive with established SMLM reconstruction methods.

%\section*{References}

\FloatBarrier
\appendix

\section{Additional experimental details}
\label{app:simulation}

\subsection{Simulation details (SMLM-C)}

\paragraph{Photophysics and localization noise.}
Each fluorophore stochastically transitions between emissive (``on'') and non-emissive (``off'') states.
On- and off-state durations are independently sampled from exponential distributions with means
$\mu_{\mathrm{on}}$ and $\mu_{\mathrm{off}}$, respectively.
All durations are measured in imaging frames.
During on periods, observed localizations are sampled from a 2D Gaussian distribution centered at the
true emitter position with standard deviation $\sigma = 10$~nm.

\paragraph{Simulation procedure.}
For each emitter, we:
(i) sample an initial on/off state according to the duty cycle;
(ii) sample on- and off-state durations from the corresponding exponential distributions;
(iii) generate noisy localization samples during on frames; and
(iv) apply a termination process depending on the imaging modality.

\paragraph{Termination models.}
For dSTORM simulations, emitters undergo irreversible photobleaching.
The total number of localization events per emitter is drawn from an exponential distribution
with mean $\mu_{\mathrm{nblink}}$.
For DNA-PAINT, we simulate both an unlimited blinking regime and a Poisson-limited regime,
in which the total number of binding events per emitter is drawn from a Poisson distribution
with mean $\lambda = 50$.

\paragraph{Detection filtering.}
To approximate limitations of standard localization pipelines, we apply a detection-limit filter
that removes localizations occurring within 500~nm of each other in the same frame.
For the experimental settings evaluated in the main paper (500$\times$500~nm region of interest),
this implies that at most one localization is retained per frame.

\paragraph{Excluded regime.}
We do not simulate the high-density, short-$t_{\mathrm{off}}$ regime.
In this regime, detection filtering removes a large fraction of localizations
due to frequent same-frame events within the 500~nm limit, resulting in heavily censored
observations that are not representative of typical reconstruction conditions.

\subsection{Evaluation metrics}
\label{app:metrics}

\paragraph{Chamfer loss.}
Let $\hat{\mathcal{P}}=\{\hat{\mathbf{x}}_i\}_{i=1}^N$ and
$\mathcal{P}=\{\mathbf{x}_j\}_{j=1}^M$ denote the predicted and ground-truth emitter sets, respectively.
The Chamfer loss is defined as
\begin{equation}
\mathcal{L}_{\text{Chamfer}} =
\frac{1}{N}\sum_{i=1}^N \min_j \lVert \hat{\mathbf{x}}_i - \mathbf{x}_j \rVert_2
+
\frac{1}{M}\sum_{j=1}^M \min_i \lVert \mathbf{x}_j - \hat{\mathbf{x}}_i \rVert_2 .
\end{equation}

\paragraph{Hungarian error.}
The Hungarian error computes the optimal bipartite matching between predicted and ground-truth
emitters using the Hungarian algorithm and reports the mean Euclidean distance of the matched pairs.
Unmatched predictions or targets are ignored.

\subsection{Computational efficiency and model size}
\label{app:compute}

 S5-S and S5-L models require approximately $6.8\,\text{h}$ and $14.1\,\text{h}$ training time respectively, whereas Mamba2-S and Mamba2-L require approximately $19.5\,\text{h}$ and $42.6\,\text{h}$. Parameter counts are: S5-S (${\sim}523\,\text{K}$), Mamba2-S (${\sim}730\,\text{K}$), S5-L (${\sim}2.08\,\text{M}$), and Mamba2-L (${\sim}2.69\,\text{M}$). 

 \clearpage
 
\subsection{Training dynamics}
\label{app:training}

\begin{figure}[h]
    \centering
    \includegraphics[width=\linewidth]{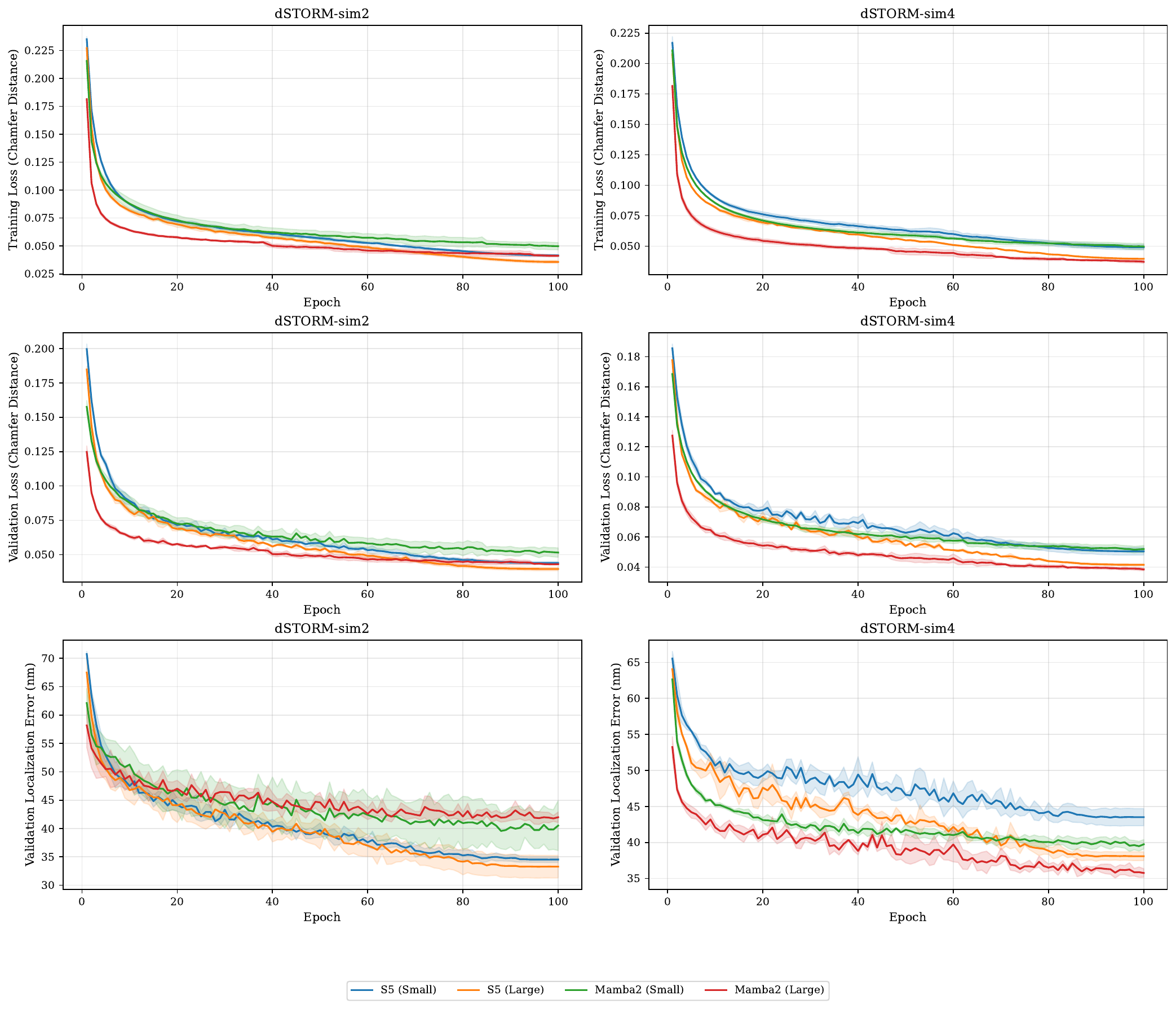}
    \caption{
    Training and validation loss curves for S5 and Mamba-2 models on dSTORM-Sim2 and dSTORM-Sim4.
    Curves are averaged over three random seeds; shaded regions indicate one standard deviation.
    Models are selected based on minimum validation localization error (Hungarian error).
    }
    \label{fig:appendix-learning-curves}
\end{figure}

Figure~\ref{fig:appendix-learning-curves} shows the training and validation loss
curves for all models on the dSTORM-Sim2 and dSTORM-Sim4 datasets.
Curves are averaged over three random seeds, with shaded regions indicating
one standard deviation. Model selection is based on minimum validation
Hungarian error.

\clearpage

\section{Additional qualitative results}
\label{app:qualitative}

\begin{figure}[h]
    \centering
    \includegraphics[width=\linewidth]{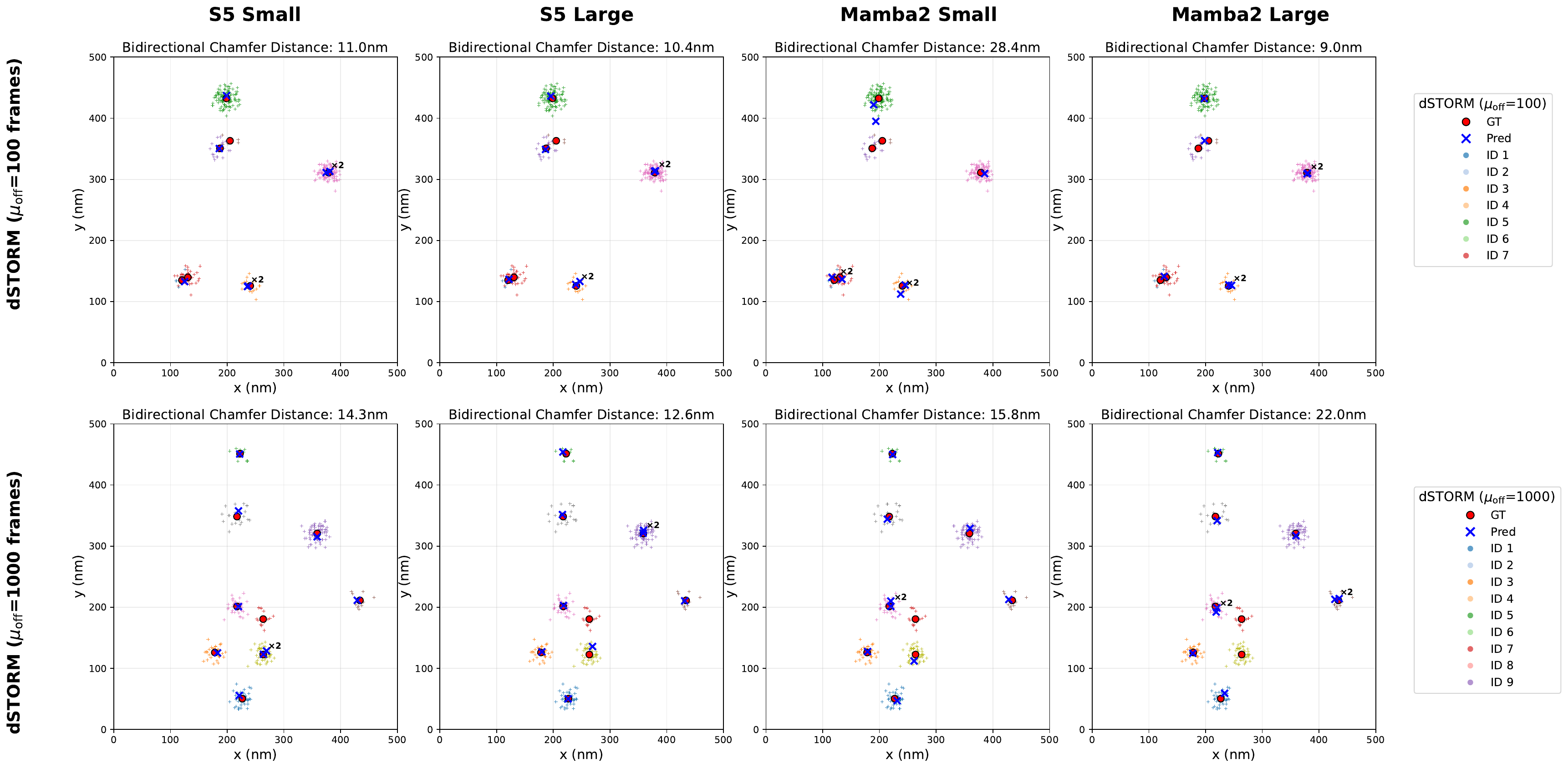}
    \caption{
    Qualitative reconstructions for representative easy test sequences (lowest 10\% of averaged Chamfer loss).
    Predictions from S5 and Mamba-2 models are shown alongside raw input frames and ground-truth emitter locations.
    }
    \label{fig:appendix-easy}
\end{figure}

\begin{figure}[h]
    \centering
    \includegraphics[width=\linewidth]{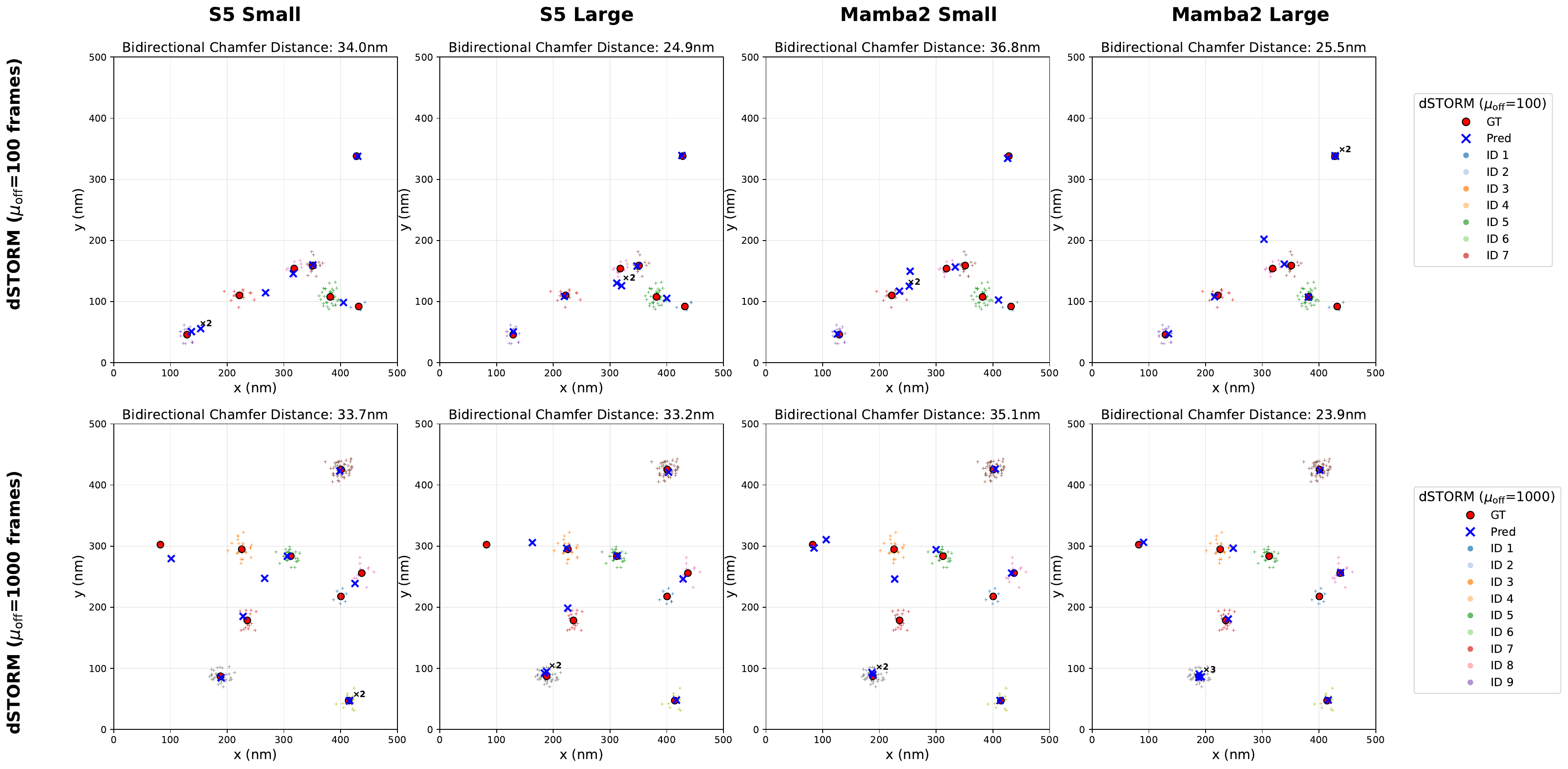}
    \caption{
    Qualitative reconstructions for representative hard test sequences (highest 5\% of averaged Chamfer loss).
    These examples highlight failure modes.
    }
    \label{fig:appendix-hard}
\end{figure}

\paragraph{Easy and hard cases.}
Using the averaged Chamfer loss across models, we identify test sequences in the lowest 10\%
and highest 5\% of the loss distribution as ``easy'' and ``hard'' cases, respectively.
Representative examples are shown in Figures~\ref{fig:appendix-easy} and \ref{fig:appendix-hard}.
These cases illustrate performance extremes and failure modes that complement
the median example presented in the main text.

\clearpage

\end{document}